\pdfoutput=1
\documentclass[11pt,letterpaper]{article}
\usepackage[margin=1in]{geometry}
\usepackage[T1]{fontenc}
\usepackage{mathptmx}
\usepackage{amsmath,amssymb}
\usepackage{graphicx}
\usepackage{booktabs}
\usepackage[table]{xcolor}
\usepackage{caption}
\usepackage{float}
\usepackage{afterpage}
\usepackage[super,sort&compress,comma]{natbib}
\usepackage[colorlinks,allcolors=biolink,breaklinks]{hyperref}
\definecolor{biolink}{rgb}{0.10,0.30,0.60}
\definecolor{oursbg}{rgb}{0.88,0.93,0.99}
\newcommand{\oursrow}{\rowcolor{oursbg}}
\newcommand{\etal}{\textit{et al.}}

\title{\textbf{How Far Can Linear Go? BiPert, a Unified Framework for\\ Single-Cell Perturbation Modelling}}
\author{Abhishek Sharma$^{1}$ and Aleksej Zelezniak$^{1,2,3}$\\[6pt]
{\normalsize $^{1}$Randall Centre for Cell \& Molecular Biophysics, King's College London, London, UK}\\
{\normalsize $^{2}$Department of Life Sciences, Chalmers University of Technology, Gothenburg, Sweden}\\
{\normalsize $^{3}$Institute of Biotechnology, Life Sciences Centre, Vilnius University, Vilnius, Lithuania}\\[4pt]
{\normalsize \href{mailto:abhishek.1.sharma@kcl.ac.uk}{abhishek.1.sharma@kcl.ac.uk}, \href{mailto:aleksej.zelezniak@kcl.ac.uk}{aleksej.zelezniak@kcl.ac.uk}}}
\date{}

\begin{document}
\maketitle
\begin{abstract}
\noindent
Understanding how genetic perturbations affect cellular behaviour is essential for studying gene function and
regulatory mechanisms. Large-scale CRISPR screens measure these effects across thousands of genes, but cover
relatively few cellular contexts. Predicting perturbation responses in unseen cell types and identifying perturbed
genes from their transcriptional effects therefore remain challenging. Here, we show that both tasks can be addressed
through a common linear decomposition of transcriptional responses into shared and perturbation-specific components.
Building on this principle, we introduce BiPert, a simple linear framework for both forward perturbation prediction
and inverse identification. For forward prediction, BiPert outperforms existing methods that use no external
biological knowledge when predicting gene expression changes in unseen cell lines. Its few-shot extension, BiPert-FS,
adapts to new cell lines using as few as ten measured perturbations. For inverse identification, BiPert-ID identifies
perturbed genes from single-cell expression profiles, outperforming deep-learning methods in unseen cell lines and
donors. We further show that perturbations whose effects transfer more consistently across cellular contexts are also
easier to identify, even after controlling for perturbation effect size. Together, these results establish a
connection between perturbation transfer and identification, and demonstrate that simple linear models provide strong,
interpretable baselines for both tasks.

\end{abstract}


\begin{figure}[tb]
\centering
\includegraphics[width=0.95\linewidth]{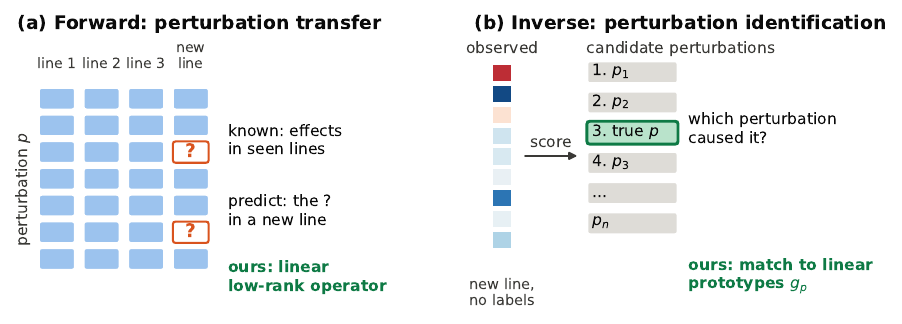}
\caption{\textbf{One linear decomposition, two directions.} Both tasks rest on one
decomposition of a perturbation's effect into a part shared across cell lines and a part specific to the
perturbation. (a) Forward: predict a perturbation's effect in a new cell line from the lines where it was
measured (Sec.~\ref{sec:task-pert}). (b) Inverse: name the perturbation behind an observed state in a new
line, with no labels from that line (Sec.~\ref{sec:task-retrieval}).}
\label{fig:tasks}
\end{figure}

\section{Introduction}
\label{sec:intro}
Large-scale perturbation screens now measure how thousands of genetic perturbations change the state of a
cell~\cite{replogle2022,nadig2025,huang2025xatlas}. Modelling these changes has two complementary directions. In the \emph{forward direction}, given the effects
of a perturbation across previously studied cell lines, we seek to predict its effect in a new cellular
context~\cite{adduri2025state,compass}. In the \emph{inverse direction}, given an observed cellular state, we seek to identify the
perturbation that most likely produced it~\cite{rnafingerprinting,expert}. The first problem enables the transfer of experimental
knowledge to unprofiled cell lines, potentially reducing the need for additional perturbation screens. The
second enables the interpretation of previously unexplained cellular states by matching them to known
perturbation responses. A key observation is that although these tasks have different prediction targets,
both require understanding which aspects of a perturbation response are shared across cellular contexts
and which are context-dependent.

Many perturbations induce a common cellular response, while retaining perturbation-specific
components~\cite{vinastorne2025systema,compass}.
Existing approaches exploit aspects of this structure. For example, COMPASS~\cite{compass} decomposes
perturbation effects into a cell-line-specific direction and a perturbation-specific residual, enabling
transfer to unseen cell lines. While COMPASS demonstrates that perturbation effects contain transferable
structure, its decomposition assumes that cellular context modifies these effects through a single shared
direction, with the remaining perturbation-specific component invariant across lines. This is a
restrictive assumption: different cellular contexts may modulate distinct components of a perturbation
response to different degrees. We therefore generalise this decomposition to a low-rank operator,
allowing each cell line to apply its own linear transformation to a shared perturbation representation.

Models designed for the forward task do not solve the inverse problem~\cite{expert}.
For inverse identification~\cite{expert,rnafingerprinting}, we exploit the same decomposition to construct
perturbation-specific prototypes from training cell lines. By removing shared cellular responses and
matching individual cells to these prototypes, we obtain a simple, training-free linear method for
identifying perturbations in unseen cell lines.

Importantly, we find that cross-cell-line transferability is not only relevant to predicting perturbation
effects, but also provides a link between the forward and inverse tasks. A transfer score estimated
entirely from training cell lines strongly predicts whether a perturbation can be identified in a
previously unseen line. Conversely, in an independent screen, we show that identifiability within a
cell type limits transfer across cell types. Together, these results suggest that a
common linear structure can support both the transfer of perturbation effects and their identification
from cellular states.

Our contributions are threefold:
\begin{enumerate}
\item \textbf{A two-way link between perturbation transfer and identification.} Transferability measured on
training cell lines alone predicts identification in held-out lines, for both BiPert-ID and the deep network;
conversely, in an independent screen, identifiability within a cell type limits transfer across cell types.
\item \textbf{Low-rank models for perturbation transfer.} We formulate cross-cell-line transfer as
structured matrix completion and introduce a low-rank operator, BiPert, generalising COMPASS, together with a
few-shot extension, BiPert-FS, for adapting to new cell lines with limited measurements.
\item \textbf{Training-free perturbation identification.} We extend the shared decomposition to inverse
identification and propose a training-free linear prototype model, BiPert-ID, that outperforms the state of
the art in unseen cell lines.
\end{enumerate}

\section{Related work}
\label{sec:related}

\paragraph{Forward: deep models for perturbation effect prediction.}
Most methods for predicting the effect of a genetic perturbation are deep networks. CPA encodes cells
into a latent space and adds a learned embedding per perturbation~\cite{lotfollahi2023cpa}; GEARS
propagates perturbations over a graph of known gene relations to reach unseen genes and
combinations~\cite{roohani2024gears}; scGPT, a transformer pre-trained on millions of cells, is fine-tuned
to the same task~\cite{cui2024scgpt}; GenePert regresses effects on gene embeddings derived from language
models~\cite{genepert}; TabICL, a tabular foundation model, predicts effects in context from external gene knowledge, six
sets of prior gene features covering protein sequence, text, morphology, protein interactions and gene
dependency~\cite{tabiclv2,palla2026tabular}; TxPert combines several biological knowledge graphs to predict effects outside the
training distribution~\cite{txpert2025}; and State, a transformer trained across many cell contexts,
targets transfer between them~\cite{adduri2025state}. Of these, only State and TxPert evaluate transfer to cell lines held out from
training.

\paragraph{Linear baselines.}
A growing body of work has questioned the effectiveness of single-cell foundation models for predicting
perturbation responses. Ahlmann-Eltze, Huber and Anders showed that a linear model fitted
to the training perturbations can match or outperform deep and foundation models on held-out perturbation
effects~\cite{ahlmannEltze2025linear}, a finding echoed by broader benchmarks of single-cell foundation
models~\cite{csendes2025benchmarking,wenteler2024perteval}. COMPASS extends this linear perspective to
cross-cell-line transfer by decomposing the effect of perturbation $p$ in cell line $\ell$ into a
cell-line-specific direction and a perturbation representation shared across cell lines (Sec.~\ref{sec:method}).
This shared representation enables prediction in unseen cell lines, outperforming GEARS, scGPT and State
in this setting~\cite{compass}. A concurrent evaluation study separates each response into a part shared
across contexts and a context-specific part, and finds that current predictors recover mainly the
former~\cite{huang2026recoverable}; it does not propose a model or study the inverse task.
PerturbMap~\cite{cui2026perturbmap} fits linear maps between treatment conditions of a single screen to
predict held-out perturbations, without holding out cell lines. We build on the
COMPASS decomposition in two directions: first, by generalising
it to a low-rank operator formulation~\cite{lock2013jive} that recovers COMPASS as a special case and adapts to a
new cell line from as few as 10 measurements; and second, by extending
the framework to the inverse problem of identifying the perturbation from the resulting cellular state.
Our few-shot model builds on subspace alignment~\cite{fernando2013subspace} and on
regularising towards a source model~\cite{evgeniou2004regularized,kuzborskij2013stability}. It is similar in spirit to geometric
matrix completion, which uses side information~\cite{boyarski2021spectral,sharma2021functional}.

\paragraph{Factor models.} Factor models decompose multiple datasets into shared and dataset-specific latent factors. JIVE~\cite{lock2013jive} separates joint and individual variation, while MOFA and MOFA+~\cite{argelaguet2018mofa,argelaguet2020mofaplus} extend factor analysis to multiple omics layers and sample groups. Guided sparse factor analysis similarly models perturbation effects through latent factors within a single CRISPR screen~\cite{zhou2023gsfa}. These methods primarily explain variation in observed data, typically requiring additional factors or loadings to be estimated for a new dataset. BiPert-FS addresses a different task: predicting unmeasured perturbation responses in a new cell line from only a few observations. Rather than decomposing the new line, it keeps the source response programs and average effects fixed, and learns only a regularised affine correction (Eq.~\eqref{eq:fewshot}). The correction is evaluated on knockdowns not used for fitting.

\textbf{Inverse: inferring the perturbation from cells.}
In contrast to predicting cellular responses given a perturbation, the inverse problem of identifying the
perturbation responsible for an observed transcriptional state has received comparatively little
attention. RNA fingerprinting~\cite{rnafingerprinting} addresses this task by matching individual cells to transcriptional
response signatures associated with known perturbations. Footprint methods such as PROGENy derive consensus response
signatures from many perturbation experiments and score new expression profiles against them to infer pathway
activity~\cite{schubert2018progeny}; BiPert-ID applies the same principle to individual gene knockdowns in single cells
of an unseen cell line. ExPert~\cite{expert},
currently the state of the art, takes a learned approach: it trains a neural encoder across multiple
perturbation screens and classifies cells by comparing their representations with learned per-gene
prototypes. In ExPert's held-out-cell-line benchmark~\cite{expert}, models designed for forward prediction (GEARS, scGPT and
GenePert) perform close to chance, showing that predicting perturbation effects does not necessarily translate into
identifying their causes.
However, none exploits the line-invariant perturbation representation introduced above.

\section{Background}
\label{sec:background}

\paragraph{Why perturbation effects decompose.} Three properties of cellular organisation motivate the
models introduced in Section~\ref{sec:method}.

\emph{Genes act in functional modules.} Proteins operate in complexes and pathways, meaning that
perturbing different members of the same module can disrupt shared cellular functions. Consistent with
this, genome-scale perturbation screens show that knockdowns of genes belonging to the same protein
complex produce strongly correlated transcriptional effects~\cite{replogle2022}. This suggests that perturbation effects occupy a low-dimensional
space of shared cellular programs.

\emph{Distinct perturbations converge on shared response directions.} Different knockdowns can engage
overlapping cellular programs and consequently produce similar transcriptional responses. Perturbations
of core growth machinery, for example, induce broad and stereotyped responses, whereas perturbations of
more specific regulators produce additional gene-dependent effects~\cite{compass}. Crucially, the strength
with which a perturbation engages this shared response direction is relatively well conserved across cell
lines, more so than the gene-specific remainder~\cite{compass}. This motivates separating a conserved perturbation component from
cell-line-dependent effects.

\emph{Cellular context determines how shared programs are expressed.} Different cell lines express
different subsets of genes and have distinct active regulatory networks. Consequently, the same
underlying cellular program can affect different downstream genes, and with different strengths, depending
on the cellular context. Perturbation responses are therefore more similar between related cell lines
than between unrelated ones~\cite{nadig2025}. This motivates modelling cell-line-specific loadings
that determine how shared programs are translated into gene-expression changes.

Together, these observations motivate the decomposition used in our models: a shared response whose
strength depends on the perturbation and the cell line, a conserved perturbation-specific component, and
a cell-line-specific weighting of shared programs.

\section{BiPert: Linear Decomposition Model}
\label{sec:method}

We consider two complementary prediction tasks corresponding to the forward and inverse directions of a
common decomposition. For each task, we provide a background formulation and introduce a linear alternative. Experiments, results and ablations follow in
Section~\ref{sec:experiments}.
Both tasks follow a cross-cell-line generalisation setting: models are trained on a set
of cell lines and evaluated on a previously unseen line. The forward task predicts the effect of a
perturbation not measured in the target line, whereas the inverse task identifies the perturbation from
the resulting cellular expression profile (Fig.~\ref{fig:tasks}). For each task, we first state the problem and
the existing approach, then introduce our linear model.

\subsection{Perturbation Transfer as Structured Matrix Completion}
\label{sec:task-pert}

\paragraph{Motivation.} Perturbation screens are available for only a limited number of cell lines, while
performing genome-scale screens across every cell type of interest is impractical. This motivates
transferring knowledge across cellular contexts: given the effects of a perturbation measured in existing
cell lines, predict its effect in a new line~\cite{compass}.

\paragraph{Task.} Let $x \in \mathbb{R}^G$ denote the expression vector of a single cell, whose entries
quantify the expression of $G$ genes. For cell line $\ell$, let $\mathcal{X}^{0}_{\ell}$ denote the set of control
cells and $\mathcal{X}_{p,\ell}$ the set of cells in which gene $p \in \{1,\dots,P\}$ has been knocked
down. We define the \emph{effect} of perturbation $p$ in line $\ell$ as the difference between the
corresponding mean expression profiles:
\begin{equation}
  y_{p,\ell} \;=\; \frac{1}{|\mathcal{X}_{p,\ell}|}\sum_{x \in \mathcal{X}_{p,\ell}} x
  \;-\; \frac{1}{|\mathcal{X}^{0}_{\ell}|}\sum_{x \in \mathcal{X}^{0}_{\ell}} x \;\in\; \mathbb{R}^G .
  \label{eq:effect}
\end{equation}
Effects $y_{p,s}$ are observed for all perturbations in the source lines $s \in \mathcal{S}$. In a target
line $\ell^{\ast} \notin \mathcal{S}$, effects are available only for a subset of perturbations
$\mathcal{O} \subset \{1,\dots,P\}$. The objective is to predict $y_{p,\ell^{\ast}}$ for every unobserved
perturbation $p \notin \mathcal{O}$. This can be viewed as matrix completion with vector-valued entries:
a $P \times L$ array, with one row per perturbation, one column for each of the $L = |\mathcal{S}| + 1$ cell
lines, and an entry $y_{p,\ell} \in \mathbb{R}^G$ in each cell, whose target column is partially observed.

COMPASS~\cite{compass} decomposes each perturbation
effect into a component along a line-specific unit vector $u_\ell \in \mathbb{R}^G$ and a residual
$g_p \in \mathbb{R}^G$ shared across cell lines:
\begin{equation}
  y_{p,\ell} \;\approx\; \beta_{p,\ell}\, u_\ell + g_p, \qquad \beta_{p,\ell} = \langle y_{p,\ell}, u_\ell \rangle .
  \label{eq:compass}
\end{equation}
Here, $\beta_{p,\ell} \in \mathbb{R}$ is a scalar coefficient and $u_\ell$ represents the dominant shared response of a cell line across perturbations, estimated as
the normalised mean perturbation effect. Biologically, this captures a common \emph{stress response}
associated with processes such as reduced cellular growth and activation of stress-response genes. Because
this component is shared across perturbations, it does not by itself distinguish which gene was perturbed;
perturbation-specific information is instead captured by the coefficient $\beta_{p,\ell}$ and the residual
$g_p$.

For a previously unseen line $\ell^{\ast}$, COMPASS-X estimates $u_{\ell^{\ast}}$ from the observed
perturbation effects $\{y_{p,\ell^{\ast}}\}_{p \in \mathcal{O}}$. It then predicts unobserved effects
$\hat y_{p,\ell^{\ast}} \in \mathbb{R}^G$ using source-line averages:
\begin{equation}
  \hat y_{p,\ell^{\ast}} \;=\; \bar\beta_p\, u_{\ell^{\ast}} + \bar g_p ,
\end{equation}
where $\bar\beta_p \in \mathbb{R}$ and $\bar g_p \in \mathbb{R}^G$ denote the averages of the corresponding
quantities across source lines.

We generalise this decomposition by representing each perturbation effect
as the product of a low-dimensional code $U_p \in \mathbb{R}^r$, shared across cell lines, and a
line-specific loading matrix $V_\ell \in \mathbb{R}^{G \times r}$, the joint structure of JIVE~\cite{lock2013jive}:
\begin{equation}
  y_{p,\ell} \;\approx\; V_\ell\, U_p .
  \label{eq:lowrank}
\end{equation}
\paragraph{BiPert.} For a new cell line $\ell^{\ast}$, we estimate its loading matrix by least squares on the observed
perturbations:
\begin{equation}
  V_{\ell^{\ast}} \;=\; \arg\min_{V \in \mathbb{R}^{G \times r}} \sum_{p \in \mathcal{O}} \lVert y_{p,\ell^{\ast}} - V U_p \rVert_2^2 .
\end{equation}
The shared codes $U_p$ are obtained from a rank-$r$ truncated SVD of the $P \times |\mathcal{S}|G$ matrix
whose row $p$ stacks the source-line effects $y_{p,s}$ across cell lines. Once $V_{\ell^{\ast}}$ is fitted, effects for unobserved perturbations are predicted as
$\hat y_{p,\ell^{\ast}} = V_{\ell^{\ast}} U_p$.

\paragraph{COMPASS-X as a special case.} COMPASS-X is recovered as a constrained instance of this
formulation. Suppose the shared residuals admit a low-rank representation $\bar g_p = W h_p$, where
$W \in \mathbb{R}^{G \times r'}$ and $h_p \in \mathbb{R}^{r'}$. Its prediction can then be written as
\begin{equation}
  \bar\beta_p u_{\ell^{\ast}} + \bar g_p \;=\;
  \begin{bmatrix} u_{\ell^{\ast}} & W \end{bmatrix}
  \begin{bmatrix} \bar\beta_p \\ h_p \end{bmatrix} .
\end{equation}
The stacked loading matrix is in $\mathbb{R}^{G \times (r'+1)}$ and the stacked code in $\mathbb{R}^{r'+1}$, so
this corresponds to Eq.~\eqref{eq:lowrank} with rank $r' + 1$, but imposes the additional constraint that
all cell lines share the same loading matrix $W$, with line-specific variation restricted to a single
direction $u_\ell$. BiPert removes this restriction, allowing each line to assign its own
weights to the shared perturbation codes.

\paragraph{BiPert-FS: our few-shot extension.} When only a small number of perturbations are observed
in the target line, fitting the full loading matrix $V_{\ell^{\ast}}$ becomes unreliable. The average response
across training cell lines provides a strong starting point for predicting a new line. Rather than learning its
response from scratch, we therefore model the target as a low-dimensional correction to this shared baseline.
Specifically, we introduce a constrained transfer model that maps the mean source-line effect,
\begin{equation*}
  \bar y_p = |\mathcal{S}|^{-1} \sum_{s \in \mathcal{S}} y_{p,s} \in \mathbb{R}^G ,
\end{equation*}
to the target line using a low-dimensional basis $\Phi \in \mathbb{R}^{G \times k}$, formed from the top singular
vectors of the source effects~\cite{fernando2013subspace}:
\begin{equation}
  \hat y_{p,\ell^{\ast}} = \underbrace{\bar y_p}_{\text{source average}}
  + \underbrace{\Phi b}_{\text{shift}}
  + \underbrace{\Phi\,(C - I_k)\,\Phi^{\top} \bar y_p}_{\text{rescale and mix programs}}.
  \label{eq:fewshot}
\end{equation}
Here, $\Phi$ has orthonormal columns and $I_k$ denotes the $k \times k$ identity matrix. The diagonal entries
of $C$ rescale individual response programs, while its off-diagonal entries capture interactions between
programs; $b$ provides an
additive shift. Thus, setting $C = I_k$ and $b = 0$ recovers the source average. We estimate
$b \in \mathbb{R}^k$ and $C \in \mathbb{R}^{k \times k}$ by ridge regression on the observed pairs
$(\bar y_p, y_{p,\ell^{\ast}})$, $p \in \mathcal{O}$, regularising $C$ towards the
identity~\cite{evgeniou2004regularized,kuzborskij2013stability} (derivation in Appendix~\ref{sec:supp-fewshot}). This
parameterisation learns an affine transformation within the low-dimensional subspace while leaving the
orthogonal component unchanged. With $k = 30$, the model has only $k + k^2 = 930$ parameters,
making it suitable for few-shot transfer.

\subsection{BiPert-ID: Perturbation Identification from Single Cells}
\label{sec:task-retrieval}

\begin{figure}[tb]
\centering
\includegraphics[width=\linewidth]{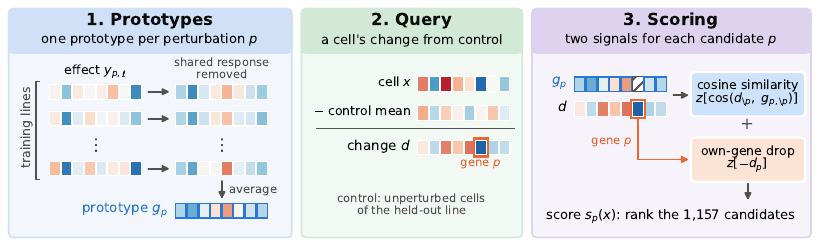}
\caption{\textbf{BiPert-ID: our training-free linear prototype model.} (1) One prototype $g_p$ per perturbation is built
from the training cell lines. (2) A query cell is represented by its change $d$ from control. (3) Each candidate $p$
is scored by cosine similarity plus own-gene drop (Eq.~\ref{eq:retrieval}).}
\label{fig:prototype}
\end{figure}

\paragraph{Motivation.} The inverse task is to identify the perturbation responsible for an observed
cellular state~\cite{expert}. In applications such as disease analysis, the cell is observed in an altered
state, but the perturbation that produced it is unknown. Differential expression captures the downstream
response but does not directly identify its cause. A perturbation atlas provides a reference of cellular
responses with known perturbations, allowing an unknown state to be matched to its likely cause. Existing
approaches construct this reference within individual screens and report limited transferability of
perturbation effects across cellular contexts~\cite{expert,rnafingerprinting}. Our decomposition addresses
this limitation by explicitly modelling cross-line transfer.

\paragraph{Task.} Given the expression vector $x \in \mathbb{R}^G$ of a perturbed cell, the goal is to
identify the knocked-down gene. We use the ExPert benchmark~\cite{expert}, which holds out entire cell
lines. Models are trained on perturbation screens from the remaining lines and evaluated on a held-out
line, which makes this a zero-shot cross-domain identification task.

\paragraph{Our training-free linear prototype model (Fig.~\ref{fig:prototype}).} Given a cell from an unseen cell line, the task is to
identify the gene that was knocked down. Our method requires no training: it constructs
perturbation-specific reference profiles from the training lines and matches the query cell against them.
\begin{enumerate}
\item \emph{Build a prototype for each perturbation.} For each knockdown $p$, measure its effect $y_{p,\ell}$
in each training cell line. Subtract the response shared by all knockdowns in that line---the general
response to perturbation---and average the residuals across lines. The result, $g_p \in \mathbb{R}^G$, is a
prototype of the gene-specific response that transfers across cell lines. It is the same residual used in
the forward-transfer model.
\item \emph{Represent the query cell.} Compare the query cell $x$ with unperturbed control cells from its own
line to obtain its expression change $d \in \mathbb{R}^G$. As in ExPert, subtract the response shared across
all perturbed query cells, leaving the change specific to the query.
\item \emph{Score each candidate gene using two signals.} First, does the query's expression change resemble
the candidate's prototype? Second, is the candidate gene itself downregulated? We combine these signals with
equal weight: prototype similarity captures the overall response pattern, while the gene's own expression
provides direct evidence of its knockdown. Candidates are ranked by
\end{enumerate}
\begin{equation}
  s_p(x) = z\!\left[\cos\!\big(d_{\setminus p},\, g_{p,\setminus p}\big)\right] + z\!\left[-d_p\right],
  \label{eq:retrieval}
\end{equation}
where $s_p(x) \in \mathbb{R}$, $\setminus p$ removes gene $p$ from both vectors, $d_p \in \mathbb{R}$ is gene
$p$'s own change, and $z[\cdot]$ standardises each signal across the 1,157 candidates. Implementation details and a score ablation are given in Appendices~\ref{sec:supp-prototype-details} and~\ref{sec:supp-ablation}.

The two tasks thus use the same decomposition in opposite directions. Transfer goes from a known gene to its
change in a new line; identification goes from an observed change in a new line to the gene whose prototype
it matches.

\section{Experiments and Results}
\label{sec:experiments}
\suppressfloats[t]
\begin{figure}[t]
\centering
\includegraphics[width=0.49\linewidth]{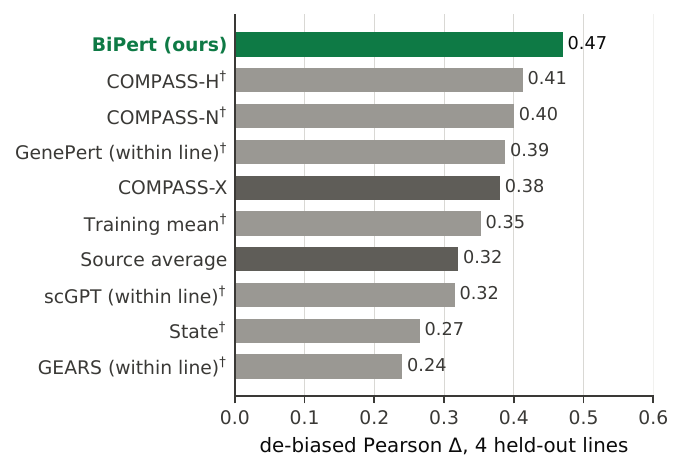}\hfill
\includegraphics[width=0.49\linewidth]{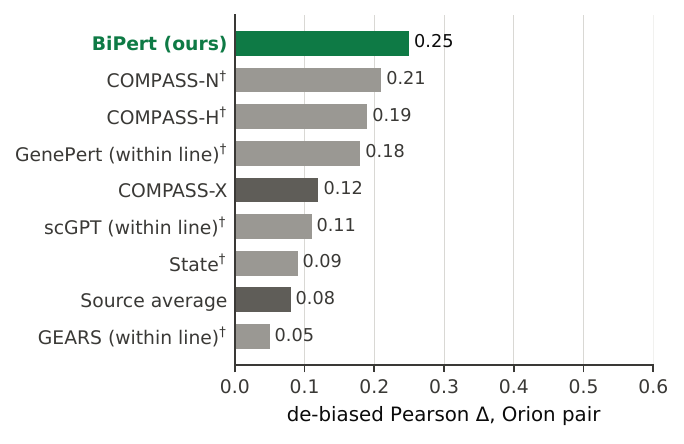}
\caption{\textbf{Mean perturbation transfer on COMPASS's benchmark}: four held-out lines (left) and the Orion
pair (right). $^{\dagger}$Published values.}
\label{fig:transfer-bars}
\end{figure}

\begin{table}[t]
\centering
\small
\begin{tabular}{@{}lcccc@{}}
\toprule
\# observed & 10 & 25 & 50 & 100 \\
\midrule
Source average & 0.320 & 0.320 & 0.320 & 0.320 \\
COMPASS-X & 0.322 & 0.355 & \textbf{0.364} & 0.370 \\
\midrule
\oursrow \textbf{BiPert (ours)} & 0.332 & 0.351 & \textbf{0.369} & \textbf{0.395} \\
\oursrow \textbf{BiPert-FS (ours)} & \textbf{0.355} & \textbf{0.361} & 0.362 & 0.366 \\
\bottomrule
\end{tabular}
\caption{\textbf{Few-shot prediction in a new line}: de-biased Pearson $\Delta$ against the number of
knockdowns observed in that line, averaged over four held-out lines and 5 splits (s.d.\ across splits
$\approx 0.012$). In this table BiPert uses rank $r = 30$.}
\label{tab:fewshot}
\end{table}

Every comparison is run on the published method's own benchmark (COMPASS~\cite{compass} or
ExPert~\cite{expert}), unless marked otherwise. We first
show perturbation transfer results, and then perturbation identification.

\textbf{Perturbation transfer.} We follow COMPASS's benchmark~\cite{compass}, which comprises CRISPR interference
(CRISPRi)~\cite{qi2013crispri} screens in four cell lines~\cite{replogle2022,nadig2025}
and in the Orion pair of cell lines~\cite{huang2025xatlas} (Appendix~\ref{sec:supp-transferbench}). We use the 2,000 most variable genes and $\log(1 + x)$ of counts normalised to $10^4$ per
cell. Each cell line is held out in turn; 20\% of its knockdowns form the test set, over five random
splits. De-biased Pearson correlation compares predicted and true effects using disjoint halves of control
cells, avoiding inflation from shared controls.

\paragraph{Implementation.} BiPert uses rank 50 (rank ablation in Appendix~\ref{sec:supp-rank}). BiPert-FS uses $k = 30$, with
ridge regularisation selected using source lines only, treating each in turn as a pseudo-target.
BiPert-ID has no free parameters. All linear models fit in seconds to minutes on a CPU.

\paragraph{Results.}\label{sec:res-pert} On COMPASS's benchmark (Fig.~\ref{fig:transfer-bars}),
BiPert predicts held-out perturbation effects more accurately than COMPASS-X across the four
cell lines (de-biased Pearson: 0.47 vs.\ 0.38) and more than doubles its accuracy on the harder Orion pair
(0.25 vs.\ 0.12). It also outperforms embedding-based regressors such as GenePert (0.39 and 0.18) and
perturbation-conditioned deep models such as State~\cite{adduri2025state}. The shared perturbation code, a central quantity in
COMPASS, is also recovered more accurately (0.34 vs.\ 0.22; Orion: 0.20 vs.\ 0.07). With few observed perturbations
(Table~\ref{tab:fewshot}), BiPert-FS leads with 10 observations and matches COMPASS-X at 25--50,
while BiPert surpasses both at 100.

\suppressfloats[t]
\begin{table}[t]
\begin{minipage}[t]{0.41\linewidth}
\centering
\small
\begin{tabular}{@{}lcc@{}}
\toprule
& Top-10 recall & MRR \\
\midrule
GEARS$^{\dagger}$ & 0.01 & 0.01 \\
scGPT$^{\dagger}$ & 0.04 & 0.02 \\
GenePert$^{\dagger}$ & 0.06 & 0.03 \\
Own-gene drop only$^{\dagger}$ & 0.18 & 0.16 \\
RNA fingerprinting$^{\dagger}$ & 0.22 & 0.14 \\
Centroid$^{\dagger}$ & 0.30 & 0.20 \\
ExPert$^{\dagger}$ & 0.540 & 0.402 \\
\midrule
\oursrow \textbf{BiPert-ID (ours)} & \textbf{0.656} & \textbf{0.521} \\
\bottomrule
\end{tabular}
\captionof{table}{\textbf{Perturbation identification} in four held-out cell lines (HepG2, Jurkat, K562, RPE1), the
lines of the transfer benchmark.
$^{\dagger}$Values released by ExPert's authors with their benchmark~\cite{expert}; means over the four lines.}
\label{tab:retrieval}
\end{minipage}\hfill
\begin{minipage}[t]{0.56\linewidth}
\centering
\small
\setlength{\tabcolsep}{4pt}
\begin{tabular}{@{}lccccc@{}}
\toprule
& & \multicolumn{2}{c}{Top-10 recall} & \multicolumn{2}{c}{MRR} \\
\cmidrule(lr){3-4}\cmidrule(l){5-6}
Held-out view & $n$ & ExPert & \begin{tabular}[c]{@{}c@{}}\textbf{BiPert-ID}\\(ours)\end{tabular} & ExPert & \begin{tabular}[c]{@{}c@{}}\textbf{BiPert-ID}\\(ours)\end{tabular} \\
\midrule
\oursrow New cell line & 10 & 0.426 & \textbf{0.539} & 0.309 & \textbf{0.415} \\
\oursrow New donor & 12 & 0.258 & \textbf{0.381} & 0.182 & \textbf{0.292} \\
New condition & 12 & \textbf{0.752} & 0.442 & \textbf{0.625} & 0.340 \\
\midrule
All views & 34 & \textbf{0.482} & 0.449 & \textbf{0.376} & 0.345 \\
\bottomrule
\end{tabular}
\captionof{table}{\textbf{Full ExPert benchmark} (22 datasets, 12 folds, 34 held-out views), on ExPert's exact test
cells; pseudobulk top-10 recall and MRR, means over views.}
\label{tab:fullbench}
\end{minipage}
\end{table}

\paragraph{Perturbation identification.} We use ExPert's full benchmark~\cite{expert}: 22 datasets in 12
folds, each holding out a cell line, a T-cell donor or a T-cell condition. The 1,157 knockdowns measured in
at least seven of ten cell contexts serve as candidates. We rebuild every dataset from public cell-level data
with ExPert's quality filters and match ExPert's released dataset statistics exactly. In each fold,
BiPert-ID is trained on the same datasets as ExPert and tested on the same cells. Following ExPert, scores are
pooled (pseudobulk): per-cell log-softmax scores are averaged over a knockdown's test cells before ranking.
We report pseudobulk top-10 recall (true perturbation among the top 10) and mean reciprocal rank (MRR),
averaged over knockdowns.

\paragraph{Results (Table~\ref{tab:retrieval}).}\label{sec:res-retrieval} ExPert~\cite{expert}, the state of the art, is a neural approach: an encoder trained
across perturbation screens embeds each cell, and candidates are scored by cosine similarity to learned
per-gene prototypes. We report results in two settings: the four held-out cell lines of the transfer benchmark, then ExPert's full
benchmark of 34 held-out views. We use these four lines because they are the lines of the transfer
benchmark, which lets Section~\ref{sec:analysis} relate both tasks on the same lines and perturbations. Besides, across all ten held-out
cell lines of the full benchmark, BiPert-ID's lead over ExPert in top-10 recall is the same as on these four
($+0.113$ vs.\ $+0.116$). On the four held-out cell lines, BiPert-ID outperforms ExPert on every line
(Table~\ref{tab:retrieval}). GEARS, scGPT and GenePert are forward models, adapted to identification by ExPert's
authors; they score close to chance (top-10 recall 0.01--0.06, against 0.009 for a random ranking). To see why, we reran GenePert and added
State~\cite{adduri2025state} (Appendix~\ref{sec:supp-forward}). Both reproduce the knockdowns and cell lines they were
trained on but not new ones, and with both, the measured effects of the training data identify more knockdowns than the
model's predictions under the same scoring.

\medskip\noindent\textit{Full benchmark (Table~\ref{tab:fullbench}).} We also evaluate ExPert's complete benchmark: 22 datasets in 12 folds, which hold out
a cell line, a T-cell donor, or a T-cell condition, scored on ExPert's exact test cells.
Averaged over all 34 views, ExPert leads slightly (pseudobulk top-10 recall 0.482
vs.\ 0.449). The gap comes entirely from views where the same donor was
seen in training under another condition. On every new-donor view and on 9 of 10 new cell lines, BiPert-ID
leads clearly: when nothing from the test line or donor was seen in training, the linear decomposition
transfers better than the deep network~\cite{expert}.

\begin{table}[b]
\small
\begin{minipage}[t]{0.47\linewidth}
\centering
\begin{tabular}{@{}lcc@{}}
\toprule
& ExPert & \textbf{BiPert-ID (ours)} \\
\midrule
Weakest-effect quarter & 0.21 & \textbf{0.31} \\
Second quarter & 0.49 & \textbf{0.66} \\
Third quarter & 0.66 & \textbf{0.77} \\
Strongest-effect quarter & 0.79 & \textbf{0.87} \\
\bottomrule
\end{tabular}
\captionof{table}{\textbf{Identification by effect size.} Pseudobulk top-10 recall in the four held-out cell lines.}
\label{tab:effect-quartile}
\end{minipage}\hfill
\begin{minipage}[t]{0.47\linewidth}
\centering
\begin{tabular}{@{}lcc@{}}
\toprule
Target gene & ExPert & \textbf{BiPert-ID (ours)} \\
\midrule
In ExPert's panel & 0.74 & \textbf{0.80} \\
Outside the panel & 0.48 & \textbf{0.61} \\
\bottomrule
\end{tabular}
\captionof{table}{\textbf{Identification by ExPert's gene panel.} Pseudobulk top-10 recall in the four held-out cell lines.}
\label{tab:panel}
\end{minipage}
\end{table}

\paragraph{Which genes are identified? (Tables~\ref{tab:effect-quartile} and~\ref{tab:panel})} An average recall does not show which knockdowns a method
finds. We therefore ask whether BiPert-ID's lead is confined to knockdowns with strong effects, and whether it comes
only from reading genes that ExPert does not see. We measure each knockdown's effect size, how strongly it
changes cells in the training cell lines, and divide the knockdowns of each held-out line into quarters, from
weakest to strongest effect. We then measure how often each method identifies the
knockdowns of each group (Table~\ref{tab:effect-quartile}). Both methods find knockdowns with strong effects
far more often than subtle ones, and BiPert-ID leads in every group.

 While the simplicity of BiPert-ID allows it to read every gene measured in the query line, ExPert uses
only a fixed panel of 3,096 genes. Even for knockdowns whose target lies inside that panel, BiPert-ID leads
(Table~\ref{tab:panel}), so its advantage is not explained solely by access to additional genes. BiPert-ID
identifies almost every knockdown that ExPert identifies (only 6\% are found by ExPert alone) and a further
18\% that ExPert misses.

\section{A Two-Way Link Between Transfer and Identification}
\label{sec:analysis}
\begin{table}[tb]
\small
\begin{minipage}[t]{0.47\linewidth}
\centering
\begin{tabular}{@{}lcc@{}}
\toprule
Transfer score (quartile) & ExPert & \textbf{BiPert-ID (ours)} \\
\midrule
Lowest  & 0.19 & 0.26 \\
Second  & 0.39 & 0.59 \\
Third   & 0.65 & 0.82 \\
Highest & 0.90 & 0.95 \\
\bottomrule
\end{tabular}
\captionof{table}{\textbf{Forward transfer predicts inverse identification.} Fraction of perturbations correctly
identified in the held-out line, stratified by within-line quartile of the
transfer score computed using training cell lines only.}
\label{tab:transfer-link}
\end{minipage}\hfill
\begin{minipage}[t]{0.47\linewidth}
\centering
\begin{tabular}{@{}lcc@{}}
\toprule
Identified within & All TFs & Strongest 10\% \\
\midrule
Both cell types & 25 / 33 & 25 / 33 \\
One cell type & 24 / 80 & 21 / 60 \\
Neither & 8 / 1,720 & 5 / 91 \\
\bottomrule
\end{tabular}
\captionof{table}{\textbf{Identifiability within a cell type limits transfer.} TFs identified across fibroblasts and
RPE-1 cells, out of those in each group.}
\label{tab:southard}
\end{minipage}
\end{table}

\paragraph{Transfer predicts identification (Table~\ref{tab:transfer-link}).} Our decomposition suggests that both tasks depend on the
same quantity: how well a perturbation's specific effect $g_p$ transfers across cell lines. We test this
using training cell lines alone. For each perturbation measured in at least two training cell lines, we calculate its
transfer score as the mean cosine similarity between its specific effect in each line and the average
across its other lines, leaving each line out in turn. We then assess whether the perturbation is
identified in a held-out line, defined as ranking among the top 10 of 1,157 candidates using pooled query
cells.

Across 3,340 perturbation--line pairs, perturbations that transfer better are identified more often: for
both methods, the share identified rises steadily from the lowest to the highest quarter of transfer
scores, with quarters cut within each held-out line. Stronger effects also transfer more readily (rank
correlation 0.54), but the link is not explained by effect size alone: at a
fixed effect size, perturbations that transfer better are still identified more often by both methods
(Appendix~\ref{sec:supp-transferlink}).
The link holds for ExPert as well, although it does not model transfer, suggesting that transferability is a useful inductive bias for identification models
in general. This association is predictive, not causal.

\paragraph{Identifiability within a cell type limits transfer (Table~\ref{tab:southard}).} We test cross-cell-type transfer in an independent screen of CRISPR
activation targeting 1,833 transcription factors (TFs) in fibroblasts and RPE-1 cells~\cite{southard2025}. For
each TF, we split its guide RNAs into two disjoint sets and test whether the effect estimated from one set
identifies the TF (top-1 among 1,833) from the other set. We perform this test within each cell type and
across cell types, considering transfer successful if either direction (fibroblast to RPE-1 or vice versa)
succeeds. Cross-cell-type identification closely tracks within-cell-type reproducibility: 25 of 33 TFs
identified independently in both cell types also transfer across them, compared with only 8 of 1,720 TFs
identified in neither. This association is not simply driven by effect strength.
Even among the 10\% of TFs with the strongest effects, cross-cell-type identification is much more frequent
for TFs identified within both cell types (25 of 33) than for those identified in neither (5 of 91). The
eight TFs reproducible within both cell types but not across them are candidates for cell-type-specific
effects. The comparison must use different guides on each side; with the same guides, it would overstate
cross-cell-type identification about fourfold (Appendix~\ref{sec:supp-guides}).

\section{Limitations and Conclusion}
\label{sec:discussion}
\textbf{Limitations.} BiPert-ID is outperformed by the deep network ExPert when the training data
contain cells of the same type as the test, as on views where the same donor was seen in training under
another condition (Table~\ref{tab:fullbench}). Methods that draw on external gene knowledge can also edge out BiPert, which uses the gene expression only:
TabICL~\cite{palla2026tabular}, given six sets of prior gene features, reaches a de-biased Pearson
correlation of 0.48 against 0.47 for BiPert, although BiPert leads on the Orion pair (0.25 vs.\
0.24)~\cite{compass}.

BiPert, a closed-form linear decomposition, matches or outperforms state-of-the-art methods on two
single-cell perturbation tasks when evaluated on held-out cell lines and donors. The learned parameters have direct
biological interpretations: perturbation codes $U_p$ reveal shared response patterns, while cell-line loadings
$V_\ell$ capture how each context reweights them. Inverse-identification scores further decompose into
prototype similarity and the knocked-down gene's own expression drop. Deep models
perform better when the test context resembles their training data, as ExPert
does on same-donor views. Benchmarks should therefore report performance on genuinely new contexts
separately from performance on closely related ones. For generalisation to new contexts, a simple linear
model is the baseline to beat.

Finally, the relationship between the two tasks persists after controlling for perturbation effect size.
This suggests that the ability to transfer perturbation-specific responses across cellular contexts is
closely linked to their identifiability from observed cells. Because the transfer score is computed using
training cell lines alone, it can also provide an advance estimate of which perturbations are likely to be
reliably identified in a new cell line, independently of the identification method used.

\section*{Data and code availability}
All datasets are public (Replogle \etal~\cite{replogle2022}, Nadig \etal~\cite{nadig2025}, X-Atlas/Orion~\cite{huang2025xatlas},
the ExPert benchmark~\cite{expert} and Southard \etal~\cite{southard2025}). The source code to reproduce the
results is available at \url{https://github.com/bach-zouk/BiPert}.

\section*{Acknowledgements}
We thank Leandro Ventimiglia for helpful discussions. This work is supported by Biotechnology and Biological Sciences Research Council (BBSRC) grant number BB/Y000730/1.
Computational resources were provided by Isambard-AI, by King's Computational Research, Engineering and Technology
Environment (CREATE) at King's College London, and by the National Academic Infrastructure for Supercomputing in
Sweden (NAISS), funded by the Swedish Research Council.

\section*{Use of AI tools}
The research direction, methodology and ideas were set by the authors. Claude Code and ChatGPT were used to assist with
scientific analysis, literature search, language editing, grammar and improving the clarity and conciseness of the
manuscript. All scientific ideas, methodology, experiments, analyses, results, and conclusions are verified by the
authors, who take full responsibility for the submitted content.

\bibliographystyle{unsrtnat}
\bibliography{refs}

\begin{thebibliography}{32}
\providecommand{\natexlab}[1]{#1}
\providecommand{\url}[1]{\texttt{#1}}
\expandafter\ifx\csname urlstyle\endcsname\relax
  \providecommand{\doi}[1]{doi: #1}\else
  \providecommand{\doi}{doi: \begingroup \urlstyle{rm}\Url}\fi

\bibitem[Replogle et~al.(2022)Replogle, Saunders, Pogson, Hussmann, Lenail,
  Guna, Mascibroda, Wagner, Adelman, Lithwick-Yanai, Iremadze, Oberstrass,
  Lipson, Bonnar, Jost, Norman, and Weissman]{replogle2022}
Joseph~M. Replogle, Reuben~A. Saunders, Angela~N. Pogson, Jeffrey~A. Hussmann,
  Alexander Lenail, Alina Guna, Lauren Mascibroda, Eric~J. Wagner, Karen
  Adelman, Gila Lithwick-Yanai, Nika Iremadze, Florian Oberstrass, Doron
  Lipson, Jessica~L. Bonnar, Marco Jost, Thomas~M. Norman, and Jonathan~S.
  Weissman.
\newblock Mapping information-rich genotype-phenotype landscapes with
  genome-scale {Perturb-seq}.
\newblock \emph{Cell}, 185\penalty0 (14):\penalty0 2559--2575.e28, 2022.
\newblock \doi{10.1016/j.cell.2022.05.013}.

\bibitem[Nadig et~al.(2025)Nadig, Replogle, Pogson, Murthy, McCarroll,
  Weissman, Robinson, and O'Connor]{nadig2025}
Ajay Nadig, Joseph~M. Replogle, Angela~N. Pogson, Mukundh Murthy, Steven~A.
  McCarroll, Jonathan~S. Weissman, Elise~B. Robinson, and Luke~J. O'Connor.
\newblock Transcriptome-wide analysis of differential expression in
  perturbation atlases.
\newblock \emph{Nature Genetics}, 57\penalty0 (5):\penalty0 1228--1237, 2025.
\newblock \doi{10.1038/s41588-025-02169-3}.

\bibitem[Huang et~al.(2025)Huang, Hsieh, Zhu, Michuda, Teng, Kim, Rumsey, Lam,
  Anigbogu, Wright, Ameen, You, Graves, Kim, Litterman, Sit, Blocker, and
  Chu]{huang2025xatlas}
Ann~C. Huang, Tsung-Han~S. Hsieh, Jiang Zhu, Jackson Michuda, Ashton Teng,
  Soohong Kim, Elizabeth~M. Rumsey, Sharon~K. Lam, Ikenna Anigbogu, Philip
  Wright, Mohamed Ameen, Kwontae You, Christopher~J. Graves, Hyunsung~John Kim,
  Adam~J. Litterman, Rene~V. Sit, Alex Blocker, and Ci~Chu.
\newblock {X-Atlas/Orion}: Genome-wide {Perturb-seq} datasets via a scalable
  fix-cryopreserve platform for training dose-dependent biological foundation
  models.
\newblock \emph{bioRxiv}, 2025.
\newblock \doi{10.1101/2025.06.11.659105}.

\bibitem[Adduri et~al.(2025)Adduri, Gautam, et~al.]{adduri2025state}
Abhinav~K. Adduri, Dhruv Gautam, et~al.
\newblock Predicting cellular responses to perturbation across diverse contexts
  with {State}.
\newblock \emph{bioRxiv}, 2025.

\bibitem[Liang and Singh(2026)]{compass}
Huan Liang and Rohit Singh.
\newblock {COMPASS}: Component-wise inference of shared and gene-specific
  perturbation response.
\newblock \emph{bioRxiv}, 2026.
\newblock doi 10.64898/2026.08.03.742643.

\bibitem[Grabski et~al.(2026)Grabski, Lee, Blair, Dalgarno, Mascio, Bradu,
  Knowles, and Satija]{rnafingerprinting}
Isabella~N. Grabski, Junsuk Lee, John~D. Blair, Carol Dalgarno, Isabella
  Mascio, Alexandra Bradu, David~A. Knowles, and Rahul Satija.
\newblock Mapping transcriptional responses to cellular perturbations with
  {RNA} fingerprinting.
\newblock \emph{Cell}, 2026.
\newblock \doi{10.1016/j.cell.2026.09.017}.

\bibitem[Schwartz et~al.(2026)Schwartz, Schlamowitz, Kuznets-Speck, Johnson,
  Das, Grody, Bai, Sun, Senthilkumar, Primavera, Jung, Kuusik, Grant, Rojas,
  Abdrabou, Peiffer, Sun, Diehl, Marr, Chandel, Mithal, Weinberg, Adli,
  Kalsotra, Prashnani, and Goyal]{expert}
Leon Schwartz, Jacob~L. Schlamowitz, Benjamin Kuznets-Speck, Conor~W. Johnson,
  Diptatanu Das, Emanuelle~I. Grody, Tao Bai, Hanxiao Sun, Irish Senthilkumar,
  Gillian Primavera, Jaekwon Jung, Britta Kuusik, Rogan~A. Grant, Enrique
  Rojas, Abdalla~M. Abdrabou, Brandon Peiffer, Zhaoli Sun, Anna~Mae Diehl,
  Carsten Marr, Navdeep~S. Chandel, Divakar~S. Mithal, Samuel~E. Weinberg,
  Mazhar Adli, Auinash Kalsotra, Ekta Prashnani, and Yogesh Goyal.
\newblock {ExPert} infers causal perturbations from cell states across
  biological and clinical contexts.
\newblock \emph{bioRxiv}, 2026.
\newblock \doi{10.64898/2026.11.02.762549}.

\bibitem[Vi{\~n}as~Torn{\'e} et~al.(2025)Vi{\~n}as~Torn{\'e}, Wiatrak, Piran,
  Fan, Jiang, Teichmann, Nitzan, and Brbi{\'c}]{vinastorne2025systema}
Ramon Vi{\~n}as~Torn{\'e}, Maciej Wiatrak, Zoe Piran, Shuyang Fan, Liangze
  Jiang, Sarah~A. Teichmann, Mor Nitzan, and Maria Brbi{\'c}.
\newblock Systema: a framework for evaluating genetic perturbation response
  prediction beyond systematic variation.
\newblock \emph{Nature Biotechnology}, 44\penalty0 (6):\penalty0 1050--1059,
  2025.
\newblock \doi{10.1038/s41587-025-02777-8}.

\bibitem[Lotfollahi et~al.(2023)Lotfollahi, Klimovskaia~Susmelj, De~Donno,
  Hetzel, Ji, Ibarra, Srivatsan, Naghipourfar, Daza, Martin, Shendure,
  McFaline-Figueroa, Boyeau, Wolf, Yakubova, G{\"u}nnemann, Trapnell,
  Lopez-Paz, and Theis]{lotfollahi2023cpa}
Mohammad Lotfollahi, Anna Klimovskaia~Susmelj, Carlo De~Donno, Leon Hetzel,
  Yuge Ji, Ignacio~L. Ibarra, Sanjay~R. Srivatsan, Mohsen Naghipourfar, Riza~M.
  Daza, Beth Martin, Jay Shendure, Jose~L. McFaline-Figueroa, Pierre Boyeau,
  F.~Alexander Wolf, Nafissa Yakubova, Stephan G{\"u}nnemann, Cole Trapnell,
  David Lopez-Paz, and Fabian~J. Theis.
\newblock Predicting cellular responses to complex perturbations in
  high-throughput screens.
\newblock \emph{Molecular Systems Biology}, 19\penalty0 (6):\penalty0 e11517,
  2023.
\newblock \doi{10.15252/msb.202211517}.

\bibitem[Roohani et~al.(2024)Roohani, Huang, and Leskovec]{roohani2024gears}
Yusuf Roohani, Kexin Huang, and Jure Leskovec.
\newblock Predicting transcriptional outcomes of novel multigene perturbations
  with {GEARS}.
\newblock \emph{Nat. Biotechnol.}, 42:\penalty0 927--935, 2024.

\bibitem[Cui et~al.(2024)Cui, Wang, Maan, Pang, Luo, Duan, and
  Wang]{cui2024scgpt}
Haotian Cui, Chloe Wang, Hassaan Maan, Kuan Pang, Fengning Luo, Nan Duan, and
  Bo~Wang.
\newblock {scGPT}: toward building a foundation model for single-cell
  multi-omics using generative {AI}.
\newblock \emph{Nature Methods}, 21\penalty0 (8):\penalty0 1470--1480, 2024.
\newblock \doi{10.1038/s41592-024-02201-0}.

\bibitem[Chen and Zou(2024)]{genepert}
Yiqun~T. Chen and James Zou.
\newblock {GenePert}: Leveraging {GenePT} embeddings for gene perturbation
  prediction.
\newblock \emph{bioRxiv}, 2024.
\newblock \doi{10.1101/2024.10.27.620513}.

\bibitem[Qu et~al.(2026)Qu, Holzm{\"u}ller, Varoquaux, and Le~Morvan]{tabiclv2}
Jingang Qu, David Holzm{\"u}ller, Ga{\"e}l Varoquaux, and Marine Le~Morvan.
\newblock {TabICLv2}: A better, faster, scalable, and open tabular foundation
  model.
\newblock \emph{arXiv preprint arXiv:2602.11139}, 2026.

\bibitem[Palla et~al.(2026)Palla, Hillsley, Kim, and Royer]{palla2026tabular}
Giovanni Palla, Alexander Hillsley, Yang-Joon Kim, and Loic~A. Royer.
\newblock Tabular foundation models are competitive cellular perturbation
  predictors across biological scales.
\newblock \emph{bioRxiv}, 2026.
\newblock \doi{10.64898/2026.06.28.735106}.

\bibitem[Wenkel et~al.(2026)Wenkel, Tu, Masschelein, Shirzad, Hodgson, Bendidi,
  Eastwood, Whitfield, Russell, El~Mesbahi, Ding, Fay, Earnshaw, Noutahi, and
  Denton]{txpert2025}
Frederik Wenkel, Wilson Tu, Cassandra Masschelein, Hamed Shirzad, Liam Hodgson,
  Ihab Bendidi, Cian Eastwood, Shawn~T. Whitfield, Craig Russell, Yassir
  El~Mesbahi, Jiarui Ding, Marta~M. Fay, Berton Earnshaw, Emmanuel Noutahi, and
  Alisandra~K. Denton.
\newblock {TxPert}: using multiple knowledge graphs for prediction of
  transcriptomic perturbation effects.
\newblock \emph{Nature Biotechnology}, 2026.
\newblock \doi{10.1038/s41587-026-03113-4}.

\bibitem[Ahlmann-Eltze et~al.(2025)Ahlmann-Eltze, Huber, and
  Anders]{ahlmannEltze2025linear}
Constantin Ahlmann-Eltze, Wolfgang Huber, and Simon Anders.
\newblock Deep-learning-based gene perturbation effect prediction does not yet
  outperform simple linear baselines.
\newblock \emph{Nature Methods}, 22\penalty0 (8):\penalty0 1657--1661, 2025.
\newblock \doi{10.1038/s41592-025-02772-6}.

\bibitem[Csendes et~al.(2025)Csendes, Sanz, Szalay, and
  Szalai]{csendes2025benchmarking}
Gerold Csendes, Gema Sanz, Krist{\'o}f~Z. Szalay, and Bence Szalai.
\newblock Benchmarking foundation cell models for post-perturbation {RNA-seq}
  prediction.
\newblock \emph{BMC Genomics}, 26\penalty0 (1):\penalty0 393, 2025.
\newblock \doi{10.1186/s12864-025-11600-2}.

\bibitem[Wenteler et~al.(2025)Wenteler, Occhetta, Branson, Curean, Huebner,
  Dee, Connell, Chung, Hawkins-Hooker, Ektefaie, Valdez~C{\'o}rdova, and
  Gallagher-Syed]{wenteler2024perteval}
Aaron Wenteler, Martina Occhetta, Nikhil Branson, Victor Curean, Magdalena
  Huebner, William Dee, William Connell, Siu~Pui Chung, Alex Hawkins-Hooker,
  Yasha Ektefaie, C{\'e}sar~Miguel Valdez~C{\'o}rdova, and Amaya
  Gallagher-Syed.
\newblock {PertEval-scFM}: Benchmarking single-cell foundation models for
  perturbation effect prediction.
\newblock \emph{bioRxiv}, 2025.
\newblock \doi{10.1101/2024.10.02.616248}.

\bibitem[Huang et~al.(2026)Huang, Wang, Li, and Wilson]{huang2026recoverable}
Yongqi Huang, Hanzhi Wang, Chenyu Li, and Parker Wilson.
\newblock The recoverable resolution of cellular perturbation-response
  prediction.
\newblock \emph{bioRxiv}, 2026.
\newblock \doi{10.64898/2026.09.30.755756}.

\bibitem[Cui et~al.(2026)Cui, Liu, and Sun]{cui2026perturbmap}
Panpan Cui, Yiqi Liu, and Wenhao Sun.
\newblock {PerturbMap}: Cross-context transfer of single-cell perturbation
  responses.
\newblock \emph{arXiv preprint arXiv:2607.28090}, 2026.
\newblock \doi{10.48550/arXiv.2607.28090}.

\bibitem[Lock et~al.(2013)Lock, Hoadley, Marron, and Nobel]{lock2013jive}
Eric~F. Lock, Katherine~A. Hoadley, J.~S. Marron, and Andrew~B. Nobel.
\newblock Joint and individual variation explained ({JIVE}) for integrated
  analysis of multiple data types.
\newblock \emph{The Annals of Applied Statistics}, 7\penalty0 (1):\penalty0
  523--542, 2013.
\newblock \doi{10.1214/12-AOAS597}.

\bibitem[Fernando et~al.(2013)Fernando, Habrard, Sebban, and
  Tuytelaars]{fernando2013subspace}
Basura Fernando, Amaury Habrard, Marc Sebban, and Tinne Tuytelaars.
\newblock Unsupervised visual domain adaptation using subspace alignment.
\newblock In \emph{ICCV}, pages 2960--2967, 2013.

\bibitem[Evgeniou and Pontil(2004)]{evgeniou2004regularized}
Theodoros Evgeniou and Massimiliano Pontil.
\newblock Regularized multi-task learning.
\newblock In \emph{KDD}, pages 109--117, 2004.

\bibitem[Kuzborskij and Orabona(2013)]{kuzborskij2013stability}
Ilja Kuzborskij and Francesco Orabona.
\newblock Stability and hypothesis transfer learning.
\newblock In \emph{ICML}, pages 942--950, 2013.

\bibitem[Boyarski et~al.(2021)Boyarski, Vedula, and
  Bronstein]{boyarski2021spectral}
Amit Boyarski, Sanketh Vedula, and Alex Bronstein.
\newblock Spectral geometric matrix completion.
\newblock In \emph{Mathematical and Scientific Machine Learning (MSML)}, 2021.
\newblock arXiv:1911.07255.

\bibitem[Sharma and Ovsjanikov(2022)]{sharma2021functional}
Abhishek Sharma and Maks Ovsjanikov.
\newblock Matrix decomposition on graphs: A functional view.
\newblock In \emph{IEEE International Conference on Acoustics, Speech and
  Signal Processing (ICASSP)}, 2022.

\bibitem[Argelaguet et~al.(2018)Argelaguet, Velten, Arnol, Dietrich, Zenz,
  Marioni, Buettner, Huber, and Stegle]{argelaguet2018mofa}
Ricard Argelaguet, Britta Velten, Damien Arnol, Sascha Dietrich, Thorsten Zenz,
  John~C. Marioni, Florian Buettner, Wolfgang Huber, and Oliver Stegle.
\newblock Multi-omics factor analysis---a framework for unsupervised
  integration of multi-omics data sets.
\newblock \emph{Molecular Systems Biology}, 14\penalty0 (6):\penalty0 e8124,
  2018.
\newblock \doi{10.15252/msb.20178124}.

\bibitem[Argelaguet et~al.(2020)Argelaguet, Arnol, Bredikhin, Deloro, Velten,
  Marioni, and Stegle]{argelaguet2020mofaplus}
Ricard Argelaguet, Damien Arnol, Danila Bredikhin, Yonatan Deloro, Britta
  Velten, John~C. Marioni, and Oliver Stegle.
\newblock {MOFA+}: a statistical framework for comprehensive integration of
  multi-modal single-cell data.
\newblock \emph{Genome Biology}, 21\penalty0 (1):\penalty0 111, 2020.
\newblock \doi{10.1186/s13059-020-02015-1}.

\bibitem[Zhou et~al.(2023)Zhou, Luo, Liang, Chen, and He]{zhou2023gsfa}
Yifan Zhou, Kaixuan Luo, Lifan Liang, Mengjie Chen, and Xin He.
\newblock A new {Bayesian} factor analysis method improves detection of genes
  and biological processes affected by perturbations in single-cell {CRISPR}
  screening.
\newblock \emph{Nature Methods}, 20\penalty0 (11):\penalty0 1693--1703, 2023.
\newblock \doi{10.1038/s41592-023-02017-4}.

\bibitem[Schubert et~al.(2018)Schubert, Klinger, Kl{\"u}nemann, Sieber, Uhlitz,
  Sauer, Garnett, Bl{\"u}thgen, and Saez-Rodriguez]{schubert2018progeny}
Michael Schubert, Bertram Klinger, Martina Kl{\"u}nemann, Anja Sieber, Florian
  Uhlitz, Sascha Sauer, Mathew~J. Garnett, Nils Bl{\"u}thgen, and Julio
  Saez-Rodriguez.
\newblock Perturbation-response genes reveal signaling footprints in cancer
  gene expression.
\newblock \emph{Nature Communications}, 9:\penalty0 20, 2018.
\newblock \doi{10.1038/s41467-017-02391-6}.

\bibitem[Qi et~al.(2013)Qi, Larson, Gilbert, Doudna, Weissman, Arkin, and
  Lim]{qi2013crispri}
Lei~S. Qi, Matthew~H. Larson, Luke~A. Gilbert, Jennifer~A. Doudna, Jonathan~S.
  Weissman, Adam~P. Arkin, and Wendell~A. Lim.
\newblock Repurposing {CRISPR} as an {RNA}-guided platform for
  sequence-specific control of gene expression.
\newblock \emph{Cell}, 152\penalty0 (5):\penalty0 1173--1183, 2013.
\newblock \doi{10.1016/j.cell.2013.02.022}.

\bibitem[Southard et~al.(2025)Southard, Ardy, Tang, O'Sullivan, Metzner,
  Guruvayurappan, and Norman]{southard2025}
Kaden~M. Southard, Rico~C. Ardy, Anran Tang, Deirdre~D. O'Sullivan, Eli
  Metzner, Karthik Guruvayurappan, and Thomas~M. Norman.
\newblock Comprehensive transcription factor perturbations recapitulate
  fibroblast transcriptional states.
\newblock \emph{Nature Genetics}, 57:\penalty0 2323--2334, 2025.
\newblock \doi{10.1038/s41588-025-02284-1}.

\end{thebibliography}

\clearpage
\appendix
\renewcommand{\thetable}{S\arabic{table}}\setcounter{table}{0}
\renewcommand{\thefigure}{S\arabic{figure}}\setcounter{figure}{0}
\renewcommand{\theequation}{S\arabic{equation}}\setcounter{equation}{0}
\section*{Supplementary Material}
\noindent The supplement is organised as follows. Appendix~\ref{sec:supp-fewshot} lists the cell lines of the transfer
benchmark, derives the few-shot fit of BiPert-FS and analyses what each of its parts contributes on real data.
Appendix~\ref{sec:supp-prototype-details} gives the implementation details of BiPert-ID. Appendix~\ref{sec:supp-ablations}
reports the ablations: the rank of BiPert and the score of BiPert-ID. Appendix~\ref{sec:supp-forward} examines why forward
models, used for identification, score low. Appendix~\ref{sec:supp-transferlink} shows that the link between transfer and
identification holds after controlling for effect size. Appendix~\ref{sec:supp-additional} gives the identification
results of the main text in more detail: the uncertainty of the comparison over ExPert's full benchmark, the results per
held-out cell line, and why the cross-cell-type analysis compares different guides.

\section{BiPert-FS}
\label{sec:supp-fewshot}
\begin{figure}[H]
\centering
\includegraphics[width=\linewidth]{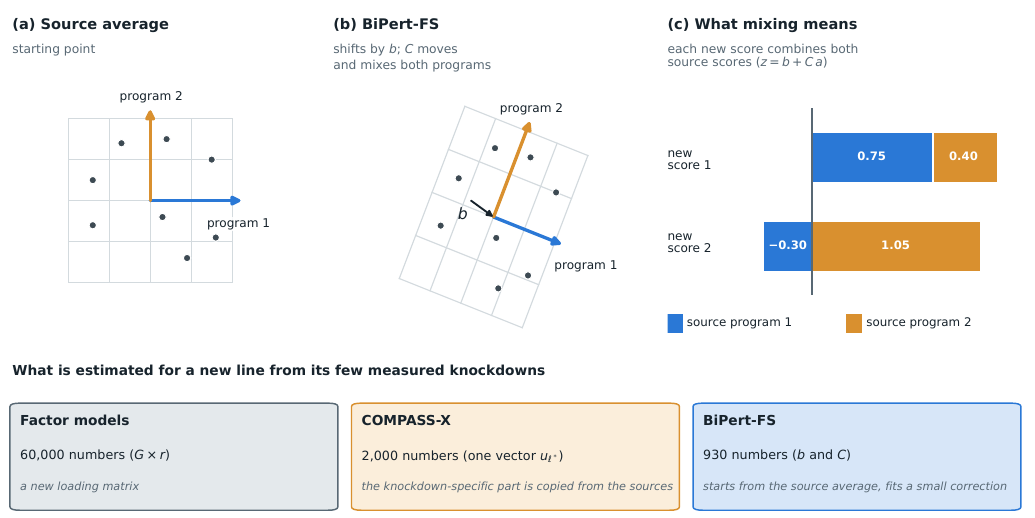}
\caption{\textbf{How BiPert-FS adapts to a new cell line.} Schematic in two dimensions.
(a) Knockdowns (dots) in the source average. (b) BiPert-FS fits an affine map $z_p \approx b + C a_p$ from source-average
scores $a_p$ to new-line scores $z_p$: $b$ shifts all knockdowns together, and $C$ moves both programs. (c) Mixing: each
new score combines both source scores. The weights on the same program (0.75, 1.05) rescale it; those on the other program
(0.40, $-0.30$) mix the two. Bottom: what each method estimates for a new line, for $G = 2{,}000$ genes, $r = 30$ factors
and $k = 30$ programs.}
\label{fig:supp-fewshot-schematic}
\end{figure}

\paragraph{Overview.} Figure~\ref{fig:supp-fewshot-schematic} illustrates the difference between BiPert-FS and
conventional factor models. Methods such as JIVE~\cite{lock2013jive} and
MOFA~\cite{argelaguet2018mofa,argelaguet2020mofaplus} jointly decompose multiple datasets into shared and dataset-specific
factors. Their primary goal is to explain the observed data, and incorporating a new dataset requires estimating its
factor loadings (Fig.~\ref{fig:supp-fewshot-schematic}, bottom).

BiPert-FS addresses a different task: predicting unmeasured perturbation responses in a new cell line from only a few
observed knockdowns. Rather than decomposing the new cell line, it retains the basis $\Phi$ learned from the source lines
and estimates a small correction, parameterised by $b$ and $C$. The matrix $C$ is regularised towards the identity, and
the correction is evaluated on knockdowns not used for adaptation.

\paragraph{Benchmark.}\label{sec:supp-transferbench}
The four cell lines of COMPASS's benchmark are K562, RPE1, HepG2 and Jurkat~\cite{replogle2022,nadig2025},
with 2,052 knockdowns measured in all four.
The Orion pair is HCT116 and HEK293T~\cite{huang2025xatlas}, with 2,270 knockdowns measured in both.

\paragraph{Inputs.} Let $\mathcal{S}$ denote the source cell lines, excluding the target line $\ell^{\ast}$. For each
source line $s$, let $Y_s$ be its knockdown-by-gene effect matrix. We construct $\Phi \in \mathbb{R}^{G \times k}$, with
$k = 30$, from the top $k$ right singular vectors of the source matrices, after centring each matrix by gene and stacking
them. The columns of $\Phi$ are orthonormal:
\[
  \Phi^{\top}\Phi = I_k .
\]
For knockdown $p$, the mean response across source lines is
\[
  \bar y_p = \frac{1}{|\mathcal{S}|}\sum_{s\in\mathcal{S}} y_{p,s} .
\]
Let $\mathcal{O}$ be the set of $n$ knockdowns observed in the target line, with responses $y_p = y_{p,\ell^{\ast}}$.

\paragraph{Low-dimensional representation.} Project the source-average and target responses onto $\Phi$:
\[
  a_p = \Phi^{\top}\bar y_p, \qquad z_p = \Phi^{\top} y_p .
\]
Both $a_p$ and $z_p$ have only $k$ coordinates. Equation~\eqref{eq:fewshot} of the main text predicts
\[
  \hat y_p = \bar y_p + \Phi\bigl(b + (C - I_k)\,a_p\bigr).
\]
Equivalently,
\[
  \hat y_p = \Phi(b + C a_p) + (I - \Phi\Phi^{\top})\,\bar y_p ,
\]
where $I$ is the $G \times G$ identity matrix. Thus, BiPert-FS adjusts the source prediction only within the learned
basis, leaving its orthogonal component unchanged.

\paragraph{Reduction to $k$ dimensions.} The prediction error separates into components inside and outside the span of
$\Phi$:
\begin{equation}
  y_p - \hat y_p = \Phi\,(z_p - b - C a_p) + (I - \Phi\Phi^{\top})(y_p - \bar y_p).
\end{equation}
Because these components are orthogonal,
\begin{equation}
\begin{aligned}
  \|y_p - \hat y_p\|_2^2 ={}& \|z_p - b - C a_p\|_2^2 \\
  &+ \|(I - \Phi\Phi^{\top})(y_p - \bar y_p)\|_2^2 .
\end{aligned}
\end{equation}
The second term does not depend on $b$ or $C$. Therefore, minimising the squared prediction error over genes reduces to
fitting a regression from $a_p$ to $z_p$ in $k$ dimensions.

\paragraph{Regularised objective.} We estimate $b$ and $C$ by minimising
\begin{equation}
  \mathcal{L}(b, C) = \sum_{p \in \mathcal{O}} \|z_p - b - C a_p\|_2^2 + \mu\,\|C - I_k\|_F^2 .
\end{equation}
The penalty encourages $C$ to remain close to the identity, while the offset $b$ is unpenalised.
Pulling $C$ towards the identity rather than towards zero means that, with little data, the model falls back on the source
programs instead of discarding them. This is needed because $C$ has $k^2 = 900$ entries, more than the $nk = 300$
coordinates observed with 10 knockdowns. The offset has only $k$ entries and its solution, $b = \bar z - C\bar a$, uses the
means of $z_p$ and $a_p$ over all observed knockdowns, so it is well determined even at small $n$ and needs no penalty. Writing $C = I_k + D$,
the objective becomes ridge regression of $z_p - a_p$ on $a_p$, with an unpenalised intercept. In other words, the model
learns how the target line differs from the source-average response. The penalty also keeps the fit well posed when fewer
than $k + 1 = 31$ knockdowns are observed.

\paragraph{Closed-form solution.} Define
\[
  X = \begin{bmatrix} 1 & a_1^{\top} \\ \vdots & \vdots \\ 1 & a_n^{\top} \end{bmatrix}, \quad
  Z = \begin{bmatrix} z_1^{\top} \\ \vdots \\ z_n^{\top} \end{bmatrix}, \quad
  \Theta = \begin{bmatrix} b^{\top} \\ C^{\top} \end{bmatrix}.
\]
Here, $X \in \mathbb{R}^{n \times (k+1)}$, $Z \in \mathbb{R}^{n \times k}$, and $\Theta \in \mathbb{R}^{(k+1) \times k}$. Let
\[
  \Lambda = \operatorname{diag}(0, \mu I_k), \qquad \Theta_0 = \begin{bmatrix} 0 \\ I_k \end{bmatrix}.
\]
The objective can then be written as
\[
  \mathcal{L}(\Theta) = \|Z - X\Theta\|_F^2 + \operatorname{tr}\bigl((\Theta - \Theta_0)^{\top} \Lambda\,(\Theta - \Theta_0)\bigr).
\]
Setting its gradient to zero gives
\[
  (X^{\top}X + \Lambda)\,\Theta = X^{\top}Z + \Lambda \Theta_0 ,
\]
and hence
\begin{equation}
  \boxed{\Theta = (X^{\top}X + \Lambda)^{-1}\,(X^{\top}Z + \Lambda \Theta_0).}
\end{equation}
This gives the few-shot adaptation parameters in closed form, without iterative training.

\paragraph{Choosing the regularisation strength.} We parameterise the penalty as
\[
  \mu = \alpha\, n\, \bar v ,
\]
where $\bar v$ is the mean variance of the source coordinates $a_p$ across knockdowns. Scaling by $n$ keeps the penalty
proportional to the data-fitting term, while $\bar v$ accounts for the scale of the perturbation effects.

We select $\alpha \in \{0.01, 0.1, 1, 10\}$ separately for each observation budget using only the source lines. In turn,
each source line is treated as a pseudo-target, with the remaining lines used to construct the source representation. We
retain the value of $\alpha$ that maximises the sum of Pearson correlation and PDS. The held-out target line is
never used for hyperparameter selection.

\begin{table}[tb]
\centering
\footnotesize
\setlength{\tabcolsep}{4pt}
\begin{tabular}{@{}lcccccccc@{}}
\toprule
& \multicolumn{4}{c}{De-biased Pearson} & \multicolumn{4}{c}{PDS} \\
\cmidrule(lr){2-5}\cmidrule(l){6-9}
 & Source avg. & Shift only & Shift + diag.\ $C$ & Full ($b$, $C$) & Source avg. & Shift only & Shift + diag.\ $C$ & Full ($b$, $C$) \\
\midrule
Mean & 0.320 & 0.353 & 0.359 & 0.366 & 0.266 & 0.258 & 0.241 & 0.261 \\
\bottomrule
\end{tabular}
\caption{\textbf{Parts of the few-shot fit} (100 observed knockdowns, mean over the four held-out lines and five splits).
\emph{Shift only}: $C = I$. \emph{Shift + diag.\ $C$}: each program rescaled separately, no mixing. \emph{Full}:
Eq.~\eqref{eq:fewshot}, the BiPert-FS row of the few-shot table.}
\label{tab:supp-fewshot-anatomy}
\end{table}

\paragraph{Contribution of each component.} To determine which parts of Eq.~\eqref{eq:fewshot} drive performance, we evaluate four
variants on the same few-shot splits and with the same regularisation selection: the source average ($b = 0$, $C = I_k$),
a shift only ($C = I_k$), a shift with per-program rescaling (diagonal $C$), and the full affine fit (Table~\ref{tab:supp-fewshot-anatomy}).

Averaged across the four cell lines, the shift accounts for most of the improvement over the source average. With 10
observed knockdowns, de-biased Pearson correlation increases from 0.320 to 0.347 with the shift alone, compared with 0.355
for the full fit. At
10 knockdowns, the shift alone explains 0.025 of BiPert-FS's 0.033 advantage over COMPASS-X. At 25 knockdowns, however,
the shift alone falls below COMPASS-X (0.351 vs.\ 0.355), and the advantage comes entirely from $C$.

The average improvement of the shift is driven by one cell line, RPE1, where it increases Pearson correlation from 0.360
to 0.517 at 100 knockdowns but reduces PDS from 0.312 to 0.141. RPE1 is the only non-cancer line of the four, which may contribute to this difference;
we did not test this. In the other three lines, the shift alone does not improve
on the source average, and the gain over the shift comes from $C$ (0.005--0.026 at 100 knockdowns). Mixing also improves
perturbation discrimination: relative to diagonal rescaling, it increases PDS in every cell line, by as much as 0.055 in
Jurkat at 100 knockdowns.

The off-diagonal entries of $C$ become reproducible across random selections of
observed knockdowns only at larger budgets. Their correlations across draws are 0.50--0.68 with 100 knockdowns, compared
with 0.03--0.22 with 10. Thus, with very few observations, $C$ primarily acts as a regularised correction close to the
identity rather than a reliably estimated program-mixing transformation. With more observed knockdowns, $C$ becomes more
effective: its gain in Pearson correlation over the shift alone rises from 0.007 at 10 knockdowns to 0.012 at 100 on
average, and from 0.012 to 0.026 in Jurkat, the line with the largest gain at 100 knockdowns.

\section{BiPert-ID: Implementation Details}
\label{sec:supp-prototype-details}
This section gives the implementation details of BiPert-ID, described in Section~\ref{sec:task-retrieval}.
\paragraph{Prototypes.} For each training cell line, perturbation effects $y_{p,\ell}$ are standardised gene-wise
(divided by a robust per-gene spread over perturbations). The shared response is removed by projecting out the
direction of the mean effect across knockdowns, and each residual is normalised to unit length. The prototype
$g_p$ is the average of these residuals over the training cell lines that measure $p$.
\paragraph{Similarity-weighted average.} Training lines are not equally relevant to a held-out line: a held-out
T-cell sample resembles the other T-cell contexts more than, for example, K562. We therefore weight each training
cell line $\ell$ by how similar its unperturbed state is to that of the held-out line $\ell^{\ast}$. Let $c_\ell, c_{\ell^{\ast}} \in
\mathbb{R}^G$ be the mean expression of their control cells ($\log(1+x)$ of counts normalised to $10^4$ per cell).
After centring each gene across $\{c_{\ell^{\ast}}\} \cup \{c_\ell\}$, we compute $\rho_\ell = \cos(c_\ell, c_{\ell^{\ast}})$, standardise
these across the training cell lines, $\tilde\rho_\ell = (\rho_\ell - \bar\rho)/\mathrm{sd}(\rho)$, and set $w_\ell = e^{\tilde\rho_\ell} /
\sum_{\ell'} e^{\tilde\rho_{\ell'}}$. The prototype becomes the weighted average
$g_p = \sum_{\ell \ni p} w_\ell\, \tilde g_{p,\ell} \big/ \sum_{\ell \ni p} w_\ell$, where $\tilde g_{p,\ell}$ is the residual above
and the sums run over the training cell lines that measure $p$. Only the held-out line's unlabelled control cells are
used, the same cells that already serve as the query reference, so the method stays zero-shot. The rule has no
tuned parameters (softmax temperature fixed at 1 in standardised units). Compared with an unweighted average over training
lines, this weighting raises pseudobulk top-10 recall by 0.014 on the four held-out cell lines and by 0.025 over all 34 views. The rule, including
the criterion for adopting it only if the 95\% confidence interval of its gain over all 34 views lay entirely above zero, was specified before evaluation.
\paragraph{Query.} A query cell's change $d$ is computed relative to the mean of the held-out line's control
cells and divided by the per-gene spread of those control cells. Following ExPert, the shared response is
removed by projecting out the direction of the mean change across all query cells of that line.
\paragraph{Scores.} The candidate's own gene is removed from both vectors before the cosine. Both terms of
Eq.~\eqref{eq:retrieval} in the main text are standardised across the 1,157 candidates and added with equal weight; there are no
tuned parameters. For a group of cells from the same knockdown, per-cell log-softmax scores are averaged
before ranking (ExPert's pseudobulk evaluation).

\section{Ablations}
\label{sec:supp-ablations}
\paragraph{BiPert: rank.}\label{sec:supp-rank}
Accuracy increases from 0.43 at rank 10 to a plateau of 0.47 at ranks 50--100, before
declining slightly to 0.44 at rank 300. The perturbation discrimination score (PDS) increases steadily with rank, from 0.15 at rank
10 to 0.25 at rank 300, matching COMPASS-X only at ranks 200--300. Thus, rank controls a trade-off between
prediction accuracy and perturbation discrimination: at rank 50, the PDS of BiPert is lower than that of
COMPASS-X (0.21 vs.\ 0.25). We also tested whether BiPert benefits from first separating components shared across all cell lines from
cell-line-specific ones, as JIVE does~\cite{lock2013jive}. It does not: replacing the plain SVD with JIVE changes accuracy by
at most 0.004 and PDS by at most 0.010 across ranks. BiPert therefore needs no such split; the improvement comes
from the low-rank structure itself.

\begin{figure}[!htb]
\centering
\includegraphics[width=0.62\linewidth]{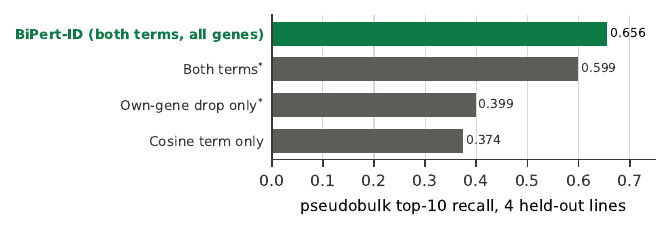}
\caption{\textbf{BiPert-ID ablation.} Mean over the four held-out lines of Table~\ref{tab:retrieval}, on ExPert's
test cells. $^{*}$Own-gene drop read only from genes measured in every training cell line.}
\label{fig:ident-ablation}
\end{figure}

\paragraph{BiPert-ID: score.}\label{sec:supp-ablation}
We ablate the scoring rule (Eq.~\eqref{eq:retrieval}) to assess the individual
importance of the cosine term and the own-gene drop term. Both are needed: either one alone identifies far fewer
perturbations than the two combined (Fig.~\ref{fig:ident-ablation}). Reading the own-gene drop from every gene
measured in the query line, rather than only from genes shared by all training cell lines, adds a further gain.

\section{Why Forward Models Identify Poorly}
\label{sec:supp-forward}
\begin{figure}[!htb]
\centering
\includegraphics[width=\linewidth]{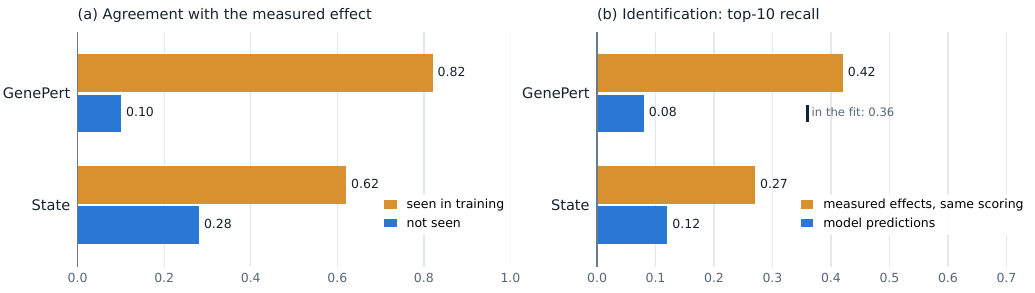}
\caption{\textbf{What forward models predict, and how well it identifies knockdowns.} (a) Correlation between the
predicted and the measured effect of a knockdown, after removing the mean effect of all knockdowns, for GenePert and State.
``Seen'' and ``not seen'' refer to knockdowns for GenePert and to cell lines for State. (b) Pseudobulk top-10 recall among
1,157 candidates on the four held-out lines, for the model's predicted profiles and for measured effects from its training
data under the same scoring. For GenePert the bar shows knockdowns left out of the fit, and the tick those included in it.
State is our own adaptation, with the authors' public zero-shot models (embedding input), and is not part of ExPert's
benchmark. The models use different training data and gene sets; compare bars only within a model.}
\label{fig:supp-forward}
\end{figure}

Having described how BiPert-ID identifies knockdowns, we now ask why models built for the forward task perform so
poorly on the same task (Table~\ref{tab:retrieval}). A forward model can be used for identification by
predicting one expression profile per candidate knockdown and matching each test cell to these profiles; the forward-model
scores in that table were released by ExPert's authors~\cite{expert}. We reran GenePert~\cite{genepert} on the four held-out cell lines to see why it scores
low, and added State~\cite{adduri2025state}, a forward model built to predict effects in an unseen cell line.
For each model we ask two questions: does the predicted profile of a knockdown agree with its measured effect once
the mean effect of all knockdowns is removed, and do measured effects, put through the same scoring, identify more
knockdowns than the predictions? Figure~\ref{fig:supp-forward} gives the numbers.

\paragraph{GenePert} fits a ridge regression from a gene's text embedding to the effect of knocking it down. For
knockdowns included in the fit, its profile is a smoothed copy of the measured effect (agreement 0.82), and matching
against these profiles identifies 36\% of the knockdowns of a held-out line, against 42\% for the measured effects
themselves. For knockdowns left out of the fit (five-fold), the agreement falls to 0.10 and identification to 8\%.
ExPert's authors report 6\% for GenePert; their adapter, as we read it, uses the fitted profiles, so we could not
reproduce that value with our re-implementation (one model per training context).

\paragraph{State} is trained on three cell lines and predicts each knockdown in the fourth from that line's control
cells. In the lines it was trained on, its predictions agree with the measured effects (0.62). In the held-out line
the agreement is 0.28, no higher than that of the plain average of the measured effects in the three training lines
(0.28), and the predicted profiles identify fewer than half as many knockdowns as that average (0.12 against 0.27).
State encodes a knockdown as a label, so it has profiles for about 920 of the 1,157 candidates.

\paragraph{Summary.} GenePert and State carry knockdown-specific signal for the knockdowns and cell lines they were
trained on, and lose it for new ones. With both models, measured effects identify more knockdowns than the model's
predictions under the same scoring. Identification in a new cell line therefore rests on the measured effects of the
candidate knockdowns, which BiPert-ID uses directly as prototypes.

\section{Transfer Predicts Identification After Controlling for Effect Size}
\label{sec:supp-transferlink}
\paragraph{Motivation.} Section~\ref{sec:analysis} and Table~\ref{tab:transfer-link} show that perturbations with higher transfer scores are more likely to be
identified in a held-out cell line. However, perturbations with larger effects may be easier to identify and may also
transfer more consistently. Indeed, transfer score and effect size have a rank correlation of 0.54. We therefore test
whether transfer predicts identification independently of effect size.

\paragraph{Setup.} We analyse 3,340 perturbation--line pairs. For each pair, the outcome is $y = 1$ if the perturbation
is identified among the top 10 of 1,157 candidates in the held-out line (pooled across cells), and $y = 0$ otherwise. We
use two predictors, both computed exclusively from training cell lines: the transfer score $t$ (defined in the main text)
and effect size $e$, measured as the logarithm of the norm of the perturbation's mean effect. Both predictors are
standardised to zero mean and unit variance.

We fit a logistic regression,
\begin{equation*}
\log \frac{P(y=1)}{P(y=0)} = \beta_t\, t + \beta_e\, e + \gamma_{\ell},
\end{equation*}
where $\gamma_\ell$ is a separate intercept for each held-out line, accounting for differences in identification
difficulty. The coefficient $\beta_t$ measures the association between transfer and identification after controlling for
effect size and cell line, and $\beta_e$ is the corresponding coefficient for effect size. A one-standard-deviation increase in transfer score multiplies the odds of identification by
$\exp(\beta_t)$. If the observed association were explained entirely by effect size, $\beta_t$ would be close to zero.

\paragraph{Results.} Table~\ref{tab:supp-transferlink} reports the fitted odds multipliers, $\exp(\beta)$. Transfer remains a strong
predictor for both methods, with 95\% confidence intervals for its odds multiplier well above 1. A one-standard-deviation
increase in transfer score multiplies the odds of identification by approximately 4.4 for ExPert and 5.2 for BiPert-ID,
compared with 1.7 and 1.9 for the same increase in effect size.

Adding transfer score to a model containing only effect size and cell-line intercepts increases McFadden's pseudo-$R^2$
from 0.16 to 0.29 for ExPert and from 0.17 to 0.30 for BiPert-ID.

As an additional check, we repeat the analysis using only the half of perturbation--line pairs with the largest effects
within each cell line. Transfer remains strongly associated with identification (odds multiplied by 4.5 for ExPert and 4.1 for
BiPert-ID), while effect size matters much less ($\times 1.2$ and $\times 1.4$). In this half, effect size and cell line alone
explain little (pseudo-$R^2$ 0.03 and 0.04), and adding transfer raises this to 0.21 and 0.19.

Together, these results show that \textbf{transfer predicts identification beyond what can be explained by perturbation
effect size alone}.

\begin{table}[tb]
\centering
\small
\begin{tabular}{@{}llccc@{}}
\toprule
& & \multicolumn{2}{c}{Odds of identification per $+1$ SD of} & Model fit (pseudo-$R^2$) \\
\cmidrule(lr){3-4}
Pairs & Method & transfer score [95\% CI] & effect size & without $\to$ with transfer \\
\midrule
All (3,340) & ExPert & $\times4.4$ $[3.8, 5.1]$ & $\times1.7$ & $0.16 \to 0.29$ \\
 & BiPert-ID & $\times5.2$ $[4.4, 6.2]$ & $\times1.9$ & $0.17 \to 0.30$ \\
Largest-effect half (1,671) & ExPert & $\times4.5$ $[3.7, 5.4]$ & $\times1.2$ & $0.03 \to 0.21$ \\
 & BiPert-ID & $\times4.1$ $[3.3, 5.1]$ & $\times1.4$ & $0.04 \to 0.19$ \\
\bottomrule
\end{tabular}
\caption{\textbf{Transfer predicts identification at fixed effect size.} Logistic regression of whether a knockdown is
identified in the held-out line (top 10 of 1,157, pooled across cells), with one intercept per line. Each multiplier is the
change in the odds of identification for a one-standard-deviation increase in that predictor, holding the other fixed;
$\times 1$ means no effect. Both predictors use training cell lines only.}
\label{tab:supp-transferlink}
\end{table}

\section{Identification Results in Detail}
\label{sec:supp-additional}
\paragraph{Full ExPert benchmark: uncertainty of the results.}\label{sec:supp-fullbench}
Across all 34 held-out views of ExPert's full benchmark (Table~\ref{tab:fullbench} and Section~\ref{sec:experiments}), ExPert achieves a mean pseudobulk top-10 recall of 0.482,
compared with 0.449 for BiPert-ID, a difference of $-0.033$ (BiPert-ID minus ExPert).

To quantify uncertainty, we use a paired bootstrap over knockdowns. Within each view, we resample knockdowns with
replacement and compute both methods' top-10 recall on the same resampled set. We then average the differences across
the 34 views, giving each view equal weight, as in ExPert's headline metric.

Over 2,000 bootstrap resamples, the 95\% confidence interval for the mean difference is $[-0.040, -0.026]$. The interval
excludes zero, indicating a small but real overall advantage for ExPert on the full benchmark.

However, this advantage comes entirely from views in which the same donor appears in training under another condition.
BiPert-ID performs better on every new-donor view and on 9 of the 10 new-cell-line views, as discussed in the main text. The corresponding intervals lie above zero
for new cell lines and new donors, and below zero for same-donor views.

\paragraph{Identification per held-out cell line.}\label{sec:supp-perline}
Table~\ref{tab:supp-perline} reports the per-cell-line results underlying the averages in Table~\ref{tab:retrieval} of the main text (Section~\ref{sec:experiments}),
evaluated on ExPert's test cells. BiPert-ID outperforms ExPert on every held-out cell line under both evaluation metrics.
For K562, we use ExPert's essential-gene screen.
\begin{table}[tb]
\centering
\small
\begin{tabular}{@{}lcccc|cccc@{}}
\toprule
& \multicolumn{4}{c|}{Top-10 recall} & \multicolumn{4}{c}{MRR} \\
& HepG2 & Jurkat & K562 & RPE1 & HepG2 & Jurkat & K562 & RPE1 \\
\midrule
GenePert$^{\dagger}$ & 0.049 & 0.063 & 0.059 & 0.084 & 0.030 & 0.031 & 0.034 & 0.040 \\
Own-gene drop only$^{\dagger}$ & 0.189 & 0.177 & 0.189 & 0.171 & 0.166 & 0.155 & 0.169 & 0.143 \\
RNA fingerprinting$^{\dagger}$ & 0.249 & 0.203 & 0.156 & 0.254 & 0.166 & 0.135 & 0.109 & 0.164 \\
Centroid$^{\dagger}$ & 0.329 & 0.314 & 0.311 & 0.255 & 0.221 & 0.210 & 0.216 & 0.166 \\
ExPert$^{\dagger}$ & 0.597 & 0.503 & 0.471 & 0.589 & 0.457 & 0.376 & 0.340 & 0.434 \\
\midrule
\oursrow \textbf{BiPert-ID (ours)} & \textbf{0.688} & \textbf{0.652} & \textbf{0.647} & \textbf{0.637} & \textbf{0.561} & \textbf{0.515} & \textbf{0.520} & \textbf{0.485} \\
\bottomrule
\end{tabular}
\caption{\textbf{Perturbation identification per held-out cell line.} $^{\dagger}$Values released by ExPert's authors with their benchmark~\cite{expert}.}
\label{tab:supp-perline}
\end{table}

\paragraph{Identifiability and transfer: why different guides are needed.}\label{sec:supp-guides}
In the CRISPR activation screen of Southard \etal~\cite{southard2025}, each transcription factor (TF) is targeted by
multiple guide RNAs. Individual guides can also produce effects unrelated to their intended TF. Consequently, comparing
the same guide across cell types may overestimate how well a TF's response transfers.

Across 1,833 TFs, the mean correlation between responses to the same guide in fibroblasts and RPE-1 cells is
0.060--0.065. This exceeds the correlation between different guides targeting the same TF within one cell type
(0.035--0.039).

The difference also affects identification. Using the same guide across cell types identifies 153--160 TFs correctly
(top-1 among 1,833 candidates), compared with only 36--46 when different guides are used. Thus, reusing guides inflates
apparent cross-cell-type identification by approximately fourfold.

To avoid this confound, the analysis in Section~\ref{sec:analysis} and Table~\ref{tab:southard} divides each TF's guides into two disjoint sets and always compares
responses obtained using different sets across cell types.

\end{document}